\documentclass{bmvc2k}
\usepackage{graphicx}
\usepackage{amsmath}
\usepackage{amssymb}

\title{TAPL: Dynamic Part-based Visual Tracking via Attention-guided Part Localization}

\addauthor{Han Wei}{whan003@e.ntu.edu.sg}{1}
\addauthor{Huang Hantao}{hantao.huang@mediatek.com}{2}
\addauthor{Yu Xiaoxi}{xiaoxi.yu@mediatek.com}{2}

\addinstitution{
 Nanyang Technological University\\
 Singapore
}
\addinstitution{
 MediaTek, Inc.\\
 Singapore
}

\runninghead{W. Han, H. Huang, X. Yu}{TAPL}

\def\etal{\emph{et al}\bmvaOneDot}

\begin{document}

\maketitle

\begin{abstract}
Holistic object representation-based trackers suffer from performance drop under large appearance change such as deformation and occlusion. In this work, we propose a dynamic part-based tracker and constantly update the target part representation to adapt to object appearance change. Moreover, we design an attention-guided part localization network to directly predict the target part locations, and determine the final bounding box with the distribution of target parts. Our proposed tracker achieves promising results on various benchmarks: VOT2018 \cite{kristan2018sixth}, OTB100 \cite{WuLY15} and GOT-10k \cite{huang2019got}. 
\end{abstract}

\section{Introduction}
Object tracking remains a challenging task due to the unconstrained real-world scenarios, such as large deformation, occlusion, illumination change and cluttered background. 
Considering aforementioned challenges, existing tracking algorithms using holistic model for object representation suffer from accuracy loss. For example, \cite{bertinetto2016fully,chen2020siamese,li2018high}  applied holistic model for object representation. Bertinetto \etal \cite{bertinetto2016fully} introduced Siamese network to efficiently match the target template and candidates in the search region by using template feature map as convolution kernel. However,  since the target representation is fixed during tracking, the tracker is susceptible to large appearance change such as deformation and occlusion. Attempts have been made to handle the issue by either exploiting local structure \cite{zhang2018structured, liang2019local} or updating the target templates online \cite{zhang2019learning, li2019gradnet}. Zhang \etal \cite{zhang2018structured} introduced a local pattern detector to identify discriminative local patterns and learns the context information among local patterns through conditional random field. 
In these recent methods \cite{zhang2018structured,liang2019local}, the local patterns are represented as peaks in feature maps, which are learned without strong semantic guidance. As a result, the detected local patterns are not associated with a clear physical meaning. 

In contrast, local patterns with clear physical meaning offer a clear explanation for tracker performance. 
Applying explicitly designed local structure in an end-to-end deep learning framework for visual tracking is still under exploration. Thus, we propose to directly localize target parts and calculate the bounding box from their distribution. An attention-guided learning strategy is introduced to provide supervision to predict more meaningful target part locations.


Moreover, to deal with large appearance change and background distractor, 
we introduce a dynamic target part representation based on the pseudo template from the most recent tracking result. A simple yet effective multi-head attention module is used to aggregate similar parts from pseudo template and template.   
Different from trackers \cite{zhang2019learning, yan2021learning} that use a large set of  historical frames, our updater only takes the ground-truth template and tracking result from the most recent frame as input and force the updater to learn to attend to the correct corresponding locations even when the prediction of the previous frame is not accurate.

The contribution of this paper can be summarized as follows: 1) We propose to formulate visual tracking as directly predicting target part locations from the target part representations. 2) The target part representations are dynamically updated based on the pseudo template generated from the most recent tracking result with our simple yet effective part representation updater. 3)  We introduce an attention loss to impose attention-guided supervision on target part localization for more reasonable part predictions. 4) Our tracker achieves comparable results to State-of-the-art works on various benchmarks: VOT2018 \cite{kristan2018sixth}, OTB100 \cite{WuLY15} and GOT-10k \cite{huang2019got}.

 
    


\section{Related Work}

\subsection{Part-based Object Model}


Part-based representation is widely used for visual tracking in traditional methods \cite{liu2015real,cehovin2013robust,li2015reliable} since it can handle large appearance change such as deformation and occlusion well. Recently, many attempts are made to exploit the local patterns in deep learning-based trackers for more robust tracking \cite{zhang2018structured, liang2019local, ma2020rpt, lukezic2020d3s, yan2021alpha}. Zhang \etal \cite{zhang2018structured} introduce a local pattern detector to identify discriminative local patterns and learn the context information among local patterns through conditional random field. Similarly, 
Liang \etal \cite{liang2019local} propose to extract local features associated with each object semantic class by integrating an auxiliary classification branch.
Lukezic \etal \cite{lukezic2020d3s} model the target and the background with pixel-level local feature vectors extracted from the target and background regions respectively. Each local feature in the search region is matched with the target and background features to generate a posterior map for segmenting foreground from background. Yan \etal \cite{yan2021alpha} adopt the pixel-wise correlation that treats each local feature in the template as a kernel to encode information of local regions in different correlation maps. The segmentation mask is predicted from the correlation maps.
In the above-mentioned methods, local parts are identified as high values in the response maps and all response maps are combined to predict the object bounding box or segmentation mask.
In contrast, our method directly predicts the location of each local part from the local representation learned with a novel guided-attention based supervision. In the later ablation study, we also show that weak supervision is the key to perform video tracking in our framework. 
\cite{lukezic2020d3s} and \cite{yan2021alpha} are excluded from our performance comparison since they focus on learning the target segmentation and requires additional segmentation dataset such as Youtube-VOS \cite{xu2018youtube} for training.
\cite{yu2020deformable} is also excluded from comparison for the same reason.

\subsection{Online Template Update}

Most trackers use a fixed template from the first frame throughout the video sequence \cite{bertinetto2016fully, li2019siamrpnpp, li2018high}.
To deal with the constantly changing object appearance, many trackers \cite{guo2017learning, zhu2018end} update the template online based on historical frames.
For example, Yang \etal \cite{MemTrack} employ a Long Short-Term
Memory (LSTM) to update the current template by encoding the previous templates in hidden state during on-line tracking.
Recent works \cite{yan2021learning,wang2021transformer} introduce transformer to directly fuse historical template features into the search region features. 
\cite{yan2021learning} uses a fixed template from the first frame and a dynamic template from the most recent previous frame. An additional classifier is trained separately to determine if the new template is reliable to update the dynamic template.  We share the same spirit as \cite{yan2021learning}, but we extract the temporal information by adopting simple yet effective dynamic target part representation.
This representation is learnt through a multi-head attention module with the input of the template and a pseudo template from the most recent tracking result. Therefore, our template update strategy is much simpler yet effective. 
Some recent works such as DROL-RPN \cite{zhou2020discriminative} and RPT \cite{ma2020rpt} propose to employ a standalone classification network, whose parameters are updated online to extract target-specific features. The online model can be integrated into an offline tracker as a complementary subnet to boost its performance. Zhou \etal \cite{zhou2020discriminative} integrate such online model to SiamRPN++ \cite{li2019siamrpnpp}. Similarly, Ma \etal \cite{ma2020rpt} apply an online model to a point-based Siamese tracker. These methods achieve good performance. However, the online learning strategy is orthogonal to our work, which can also be used to boost our performance.  Thus, the methods boosted by the online discriminative model are not considered in our performance comparison.

\subsection{Transformer Based Visual Tracking}
Transformer \cite{transformer} is the state-of-the-art model for language tasks because of its capability to capture global dependencies among all inputs. It has been applied for object detection in DETR \cite{carion2020end} and achieved comparable performance to CNN based detectors. Inspired by DETR, recent works \cite{chen2021transformer,yan2021learning,wang2021transformer} adopt transformer for visual tracking.
Yan \etal \cite{yan2021learning} input both template and search region features to transformer encoder to model the spatial temporal feature dependencies, while the decoder learns a query embedding. The resulting feature map of the search region is used to predict the corner points of object with a fully-convolutional head. Our tracker differs from existing transformer-based trackers as: existing methods represent the target object as a global template feature map and predict target bounding box from the search region feature map fused with template features. In contrast, we treat elements in the template feature map as independent parts and predict their location separately. The bounding box is estimated based on the distribution of the predicted parts coordinates.


\section{Methodology}

Different from existing works \cite{yan2021learning, yan2021learning}, we formulate visual tracking as a two-stage problem, which we first directly predict target part locations and then estimate the bounding box from their distribution in an end-to-end deep framework. The key components of the proposed framework are the dynamic target part representation module and the attention-guided part localization module, which will be elaborated in later part of this section.

\begin{figure}
    \centering
    \includegraphics[width=0.9\textwidth]{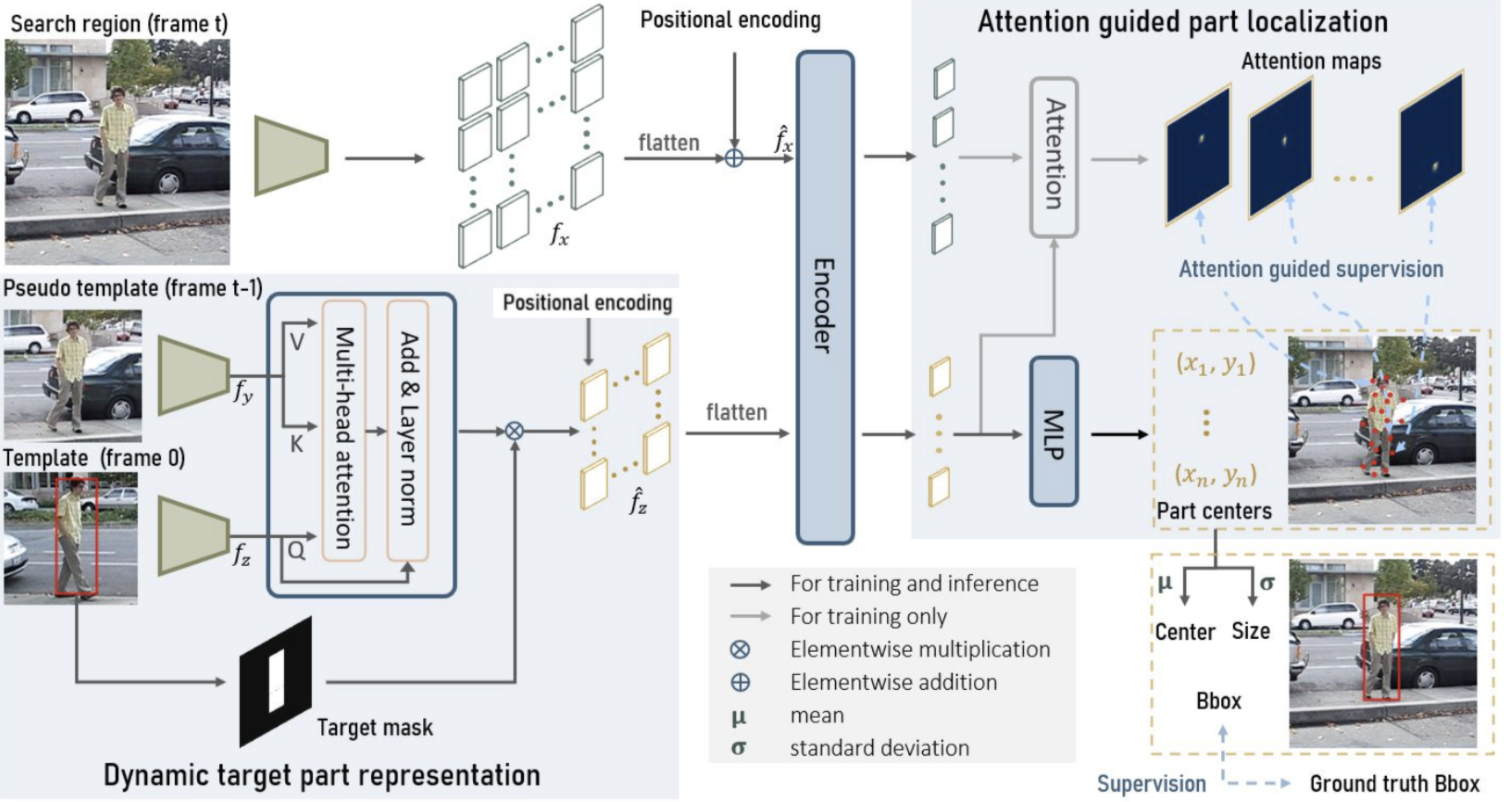}
 \caption{Overview of the proposed framework. Our tracker takes the search region, template and a pseudo template as inputs. The target object is represented as part vectors and dynamically updated with the most recent pseudo template in the dynamic target part representation module. The target parts and search region parts are input to the transformer encoder to encode global information. Encoder output of target parts are used to directly predict the location of the corresponding part with the attention-guided supervision. The bounding box is calculated with the part distribution and supervised by the ground truth.    }\label{fig.arch}
\end{figure}

As illustrated in Figure \ref{fig.arch}, our network takes an image triplet as input (i.e., a template image $z$ cropped from the initial frame, a search region $x$ from the current frame and the pseudo template $y$ cropped from the tracking result in the previous frame) and feed them through the same backbone convolutional network to extract generalized high-level features. The resulting lower-resolution feature maps obtained from $x$, $y$ and $z$ are denoted as  $f_x \in \mathbb{R}^{C \times H_x \times W_x}$, $f_y \in \mathbb{R}^{C \times H_y \times W_y}$ and $f_z \in \mathbb{R}^{C \times H_z \times W_z}$ respectively. Each location in the feature maps are used as local part representation. The target part representations are dynamically updated with the most recent pseudo template in the dynamic target part representation module. We use a transformer encoder to globally encode information from all parts. Encoder output embedding of target parts are used to directly predict their locations in the search region with the attention-guided supervision. Finally, the bounding box is calculated with the part distribution. 

\subsection{Dynamic Target Part Representation}
The local target part representation defined as $H_z \cdot W_z$ number $C$-dimensional vectors from the template feature map 
$f_z \in \mathbb{R}^{C \times H_z \times W_z}$. Each local part representation corresponds to a local region in the input template image. 
However, $f_z$ contains both target parts and background parts because all template images are cropped into a fixed square region which does not precisely bound the object in the data preparation process. To remove the effect of backgrounds, we introduce a target mask $\mathcal{M}_z \in \mathbb{R}^{1 \times H_z \times W_z}$ to zero out the background parts in the feature maps. To generate the mask, we first calculate the part centers $\{c_i\in \mathbb{R}^2\}_{i=1}^{H_z \cdot W_z}$ mapped to the scale of template $z$. The part centers located inside the ground truth bounding box are considered target parts. The mask values corresponding to the target parts are set to 1 while others are set to 0. As shown in Fig. \ref{fig.arch}, the target part representation will be further updated by a multi-head attention module.

\paragraph{Attention-based Part Representation Updater}
The appearance of the target object constantly changes during tracking, thus its crucial to dynamically update the target representation with its most recent appearance. To this end, we propose a simple yet effective template feature updater based on multi-head attention \cite{transformer} to conduct part-wise updating only using the most recent one pseudo template.
The incentive of the updater is to align each part in the pseudo template to its most attentive target part in the template and fuse their representation vectors. A learnable positional vector is also added in the dynamic part representation to encode geometric information of the target parts.
The dynamic part representations $\{ \hat{f_z}(i) \in \mathbb{R}^{1\times C} , i = 1,...,H_z\cdot W_z \}$ are calculated as:
\begin{equation}
\begin{aligned} 
    & \hat{f_z}(i) = f_z(i) + \text{Pos}_z(i)+ \text{Atten}(Q, K)V, \quad  \forall \quad i=1,...,H_z \cdot W_z , \\
   &  K=V=[f_y(1),...,f_y(H_y\cdot W_y)] \in \mathbb{R}^{(H_y \cdot W_y) \times C}, \\
   & Q=f_z(i)  \in \mathbb{R}^{1\times C} \
\end{aligned}
\end{equation}
where Q, K and V denote the query, key, and value vectors for the multi-head attention module $\text{Atten()}$ as introduced in \cite{transformer}. $\text{Pos}_z(i) \in \mathbb{R}^{1\times C}$ denotes the learnable positional encoding vector at position $i$ similar to that used in DETR \cite{carion2020end}.

Notably, since the reliability of the pseudo template generated from the tracking results is not guaranteed, we use template parts as query and fuse pseudo template part features onto their representations, in such a way a misaligned pseudo template can be recovered.
During training, to simulate the uncertainty in the tracking result, we randomly shift and scale the ground truth region to generate pseudo template $y$.

\subsection{Attention-guided Part Localization}

We aim to predict target part locations directly from their representation vectors. To this end, we adopt the transformer encoder to encode the part representations with positional information of the search region and achieve global information exchange among parts, 
\paragraph{Global Feature Encoding}
In addition to the target part representation, each feature vector of the search region can also be regarded as a local part. Due to the permutation invariance of the transformer input, we add sinusoidal positional embeddings to the search region feature vectors following \cite{carion2020end}:
\begin{equation}
  \hat{f_x}(i) = f_x(i) + \text{Pos}_x(i), \quad i = 1,...,H_x\cdot W_x
\end{equation}
where $\text{Pos}_x(i)$ denotes the sinusoidal positional embedding for spatial position $i$.
Additionally, we apply the target mask $\mathcal{M}_z$ on the updated template feature map $\hat{f}_z$ to ensure only the target parts are considered in the feature embedding process.
Then the feature vectors of search region parts $\{ \hat{f_x}(i)\in \mathbb{R}^{1\times C}, i = 1,...,H_x \cdot W_x \} $ and dynamic target parts $\{ \hat{f_z}(i) \in \mathbb{R}^{1\times C} , i = 1,...,H_z \cdot W_z \}$ are concatenated to form a sequence of $H_x \cdot W_x+ H_z \cdot W_z$ part feature vectors of dimension $C$ as input to the the stacked transformer encoder. 
The encoder layers enable each part representation to globally encode information from all other parts via the multi-head self-attention module. The encoded part representations then go through a residual connection and a feed-forward network to generate the globally encoded part representations.

\paragraph{Attention-guided Part Localization }
The transformer encoder output the encoded part representation vector for each input part vector of both the search region and template, which are denoted as $\{h_x(i)  \in \mathbb{R}^C, i = 1,...,H_x \cdot W_x \}$ and $\{h_z(i)  \in \mathbb{R}^C, i = 1,...,H_z\cdot W_z \}$, respectively. We employ a multi-layer perceptron (MLP) consisting of two linear layers for part localization. Each part representation vector in the template $\{h_z(i)  \in \mathbb{R}^C, i = 1,...,H_z\cdot W_z \}$ is input to the MLP to predict its relative position $l_i \in \mathbb{R}^{1\times 2}$ in the search region as:
\begin{equation}
  l_i = \text{MLP}(h_z(i)), \quad i = 1,...,H_z \cdot W_z
\end{equation}

We estimate the target bounding box center $o$ and scale $s$ based on the target parts distributions as:
\begin{equation}
\begin{aligned}
       o = \frac{1}{N_t} \sum_{i=1}^{H_z\cdot W_z} l_i\mathcal{M}_z(i), \ \   s = \frac{\sigma}{N_t} \sqrt{\sum_{i=1}^{H_z\cdot W_z} (l_i-o)^2\mathcal{M}_z(i)}
\end{aligned}
\end{equation}
where $N_t$ is the number of target parts in the template feature, which is calculated by $N_t=\sum_i M_z(i)$ and $\mathcal{M}_z(i) \in \{0, 1\}$ is the mask value corresponding to part $f_z(i)$. $\sigma$ is a scaling parameter fixed at 3, which is calculated with the assumption that all target parts are uniformly distributed in the bounding box. We penalize the bounding box prediction by a combination of a L1 loss and a generalized IoU loss  \cite{Rezatofighi_2018_CVPR} between the estimated bounding box and ground truth provided. The combined loss is denoted as $\mathcal{L}_{bbox}$


An obvious challenge for part-based tracking is the part level label, which is difficult to get. Bounding box label can not directly provide part level information. To tackle this challenge, we propose a novel attention loss to guide the part localization in a self-supervised fashion. The incentive behind the attention loss is to force a predicted part location to agree with its most attentive location in the search region according to the attention map. 
The attention vector $a_i \in \mathbb{R}^{1 \times (H_x\cdot W_x)}$ from a target part $h_z(i)$ to the search region parts is calculated as:
\begin{equation}
  a_i =   \text{Gumbel}(h_z(i)[h_x(1),...,h_x(H_x \cdot W_x)]^T)
\end{equation}
where Gumbel() is the gumbel-softmax function \cite{gumbel} that generates a differentiable approximation of the hard attention vector, i.e., an one-hot vector where value 1 corresponds to the most attentive location.
The attention loss is computed as the summation of the L1 distance between the attended part locations and the predicted part locations :
\begin{equation}
\begin{aligned}
  \mathcal{L}_{atten} =   
 \sum_{i=1}^{H_z\cdot W_z} || a_i P^T - l_i||_1
\end{aligned}
\end{equation}
 $P=[p_1,...,p_{H_x\cdot W_x}] \in \mathbb{R}^{2\times (H_x\cdot W_x)}$ contains the x and y coordinates of all search region parts. 
 The total loss function is calculated as:
 \begin{equation}
\begin{aligned}
  \mathcal{L} =  \mathcal{L}_{bbox} + \lambda \mathcal{L}_{atten}
\end{aligned}
\end{equation}
 where $\lambda$ is a balancing factor empirically set to 0.1.
 This proposed loss design makes attention maps as an explicit guidance for the part location prediction and offers clear physical meaning for each part prediction. Moreover, the use of gumbel-softmax enables our network to generate a differentiable hard attention, which can be used in end-to-end learning. As a result, during training both the attention and part location predictions are learnable. Since the loss $\mathcal{L}_{bbox}$ constrains the overall distribution of predicted part locations, it indirectly imposes supervision to attention learning through the attention loss.
 With synergy from the attention-guided part localization, the final network performance is further improved. 



\section{Experiments}
We evaluate our tracker on VOT2018 \cite{kristan2018sixth}, OTB100 \cite{WuLY15} and GOT-10k \cite{huang2019got} benchmarks.
The model is trained on COCO \cite{lin2014microsoft}, ImageNet
VID \cite{deng2009imagenet}, LaSOT \cite{fan2019lasot} and GOT-10k \cite{huang2019got}. Notably, we only use GOT-10k training set for evaluation on GOT-10k following the requirements of the benchmark. Each training sample consists of a search region, a template and a pseudo template. The search region and template are randomly selected from two frames within a range of 100 frames in a video sequence. The pseudo template is set as the previous frame of the search frame.

\subsection{Implementation Details}
The proposed tracker is implemented in Pytorch and trained on 4 RTX-2080Ti GPUs. A ResNet-50 pretrained on ImageNet \cite{deng2009imagenet} is used as the backbone for feature extraction. Feature maps from its last three convolutional layers are concatenated and down-sampled to a dimension of 512 channels. In the template updater and transformer encoder, the number of heads and hidden dimension of the multi-head self attention layers are set to 8 and 512 respectively. We stack 4 encoder layers to form transformer encoder. We train the network for 40 epochs with stochastic gradient descent (SGD). The first 5 epochs are warm-up period with a linearly increased learning rate from 0.001 to 0.005. Afterwards, the learning rate will exponentially decay to 0.0005 till the end. The backbone parameters from the last three layers are trained after the first 10 epochs, while other parameters of the backbone are fixed. 
\subsection{State-of-the-art Comparison}

\paragraph{VOT2018}
VOT2018 dataset contains 60 challenging video sequences. We evaluate our performance using the expected average overlap (EAO) \cite{kristan2018sixth}, which considers both accuracy and robustness of the tracker.
We demonstrate the accuracy score (A), robustness score (R) and the EAO score of our tracker and recently proposed State-of-the-art methods including SiamRPN++ \cite{li2019siamrpnpp}, DiMP\cite{bhat2019learning}, Ocean \cite{zhang2020ocean}, SiamBAN \cite{chen2020siamese}, KYS \cite{bhat2020know}, TrDiMP \cite{wang2021transformer}, SiamRN \cite{cheng2021learning} in Table \ref{tab:vot},

\begin{table}[h]
\centering
\scalebox{0.68}{
\begin{tabular}{l c c c c c c c c c}
\hline
   & SiamRPN++ \cite{li2019siamrpnpp} & DiMP\cite{bhat2019learning} & Ocean \cite{zhang2020ocean} & SiamBAN \cite{chen2020siamese} & KYS \cite{bhat2020know} &  TrDiMP \cite{wang2021transformer} & SiamRN \cite{cheng2021learning}  & \textbf{Our} \\ 
    & \texttt{CVPR2019}  & \texttt{ICCV2019} & \texttt{ECCV2020} & \texttt{CVPR2020}  & \texttt{ECCV2020}  & \texttt{CVPR2021} & \texttt{CVPR2021} &  \\ 
   \hline
\textbf{A}($\uparrow$)          & \textcolor{green}{0.604}  & 0.597 &  0.598 & 0.597  & \textcolor{blue}{0.609} & 0.600 & 0.595 & \textcolor{red}{0.617}\\ 
\textbf{R}($\downarrow$)     & 0.234  &  0.152 &  0.169  & 0.178 & 0.143 & \textcolor{green}{0.141} & \textcolor{red}{0.131} & \textcolor{blue}{0.140}\\ 
\textbf{EAO}($\uparrow$)    & 0.417  &  0.441 & \textcolor{green}{0.467} & 0.452 & 0.462 & 0.462 & \textcolor{blue}{0.470} & \textcolor{red}{0.489}\\ \hline
\end{tabular}
}
\vspace*{0.5mm}
\caption{\label{tab:vot} Comparison on VOT2018 with the state-of-the-art in terms of EAO score, acccuracy score (A) and robustness score (R). The top-3 results are
shown in red, blue and green, respectively. DiMP denotes the ResNet-50 version (DiMP-50), and Ocean denotes the offline version.
}
\end{table}

Based on Table \ref{tab:vot}, our tracker achieves the best EAO score, which shows an improvement of 4\% from 0.470 to 0.489. Notably, our accuracy score also outperforms other methods by a large margin, which implies that it can obtain more accurate bounding box under object deformation. This may be attributed to our more flexible part representation.

\begin{figure}
    \centering
    \subfigure[]{\includegraphics[width=0.4\textwidth]{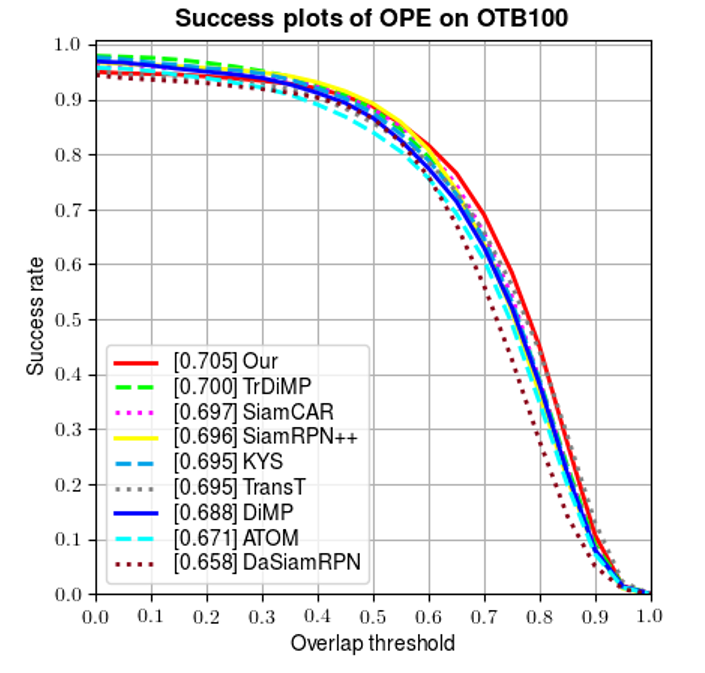}} 
    \subfigure[]{\includegraphics[width=0.4\textwidth]{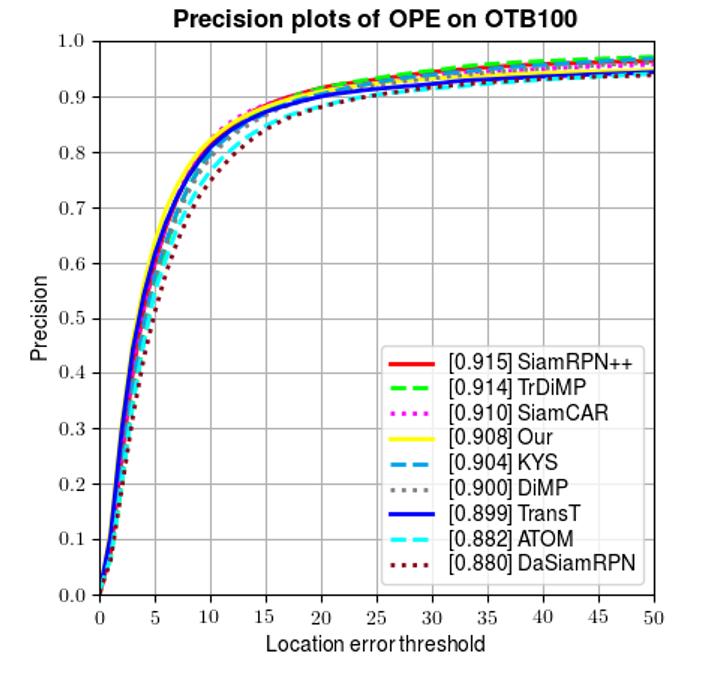}} 
    \caption{(a) Success plots and (b) precision plots of our tracker and state-of-the-art methods on OTB100 benchmark. The TrDiMP result shown is generated from the raw result file provided by the authors, which is slightly different from the reported result}
    \label{fig:otb}
\end{figure}

\paragraph{OTB100}
OTB100 is a classic benchmark for visual tracking, which consists of 100 sequences. Following the protocols of the benchmark \cite{WuLY15}, we evaluate our method against the recent state-of-the-arts with the success plot and precision plot as shown in Figure \ref{fig:otb}. The compared state-of-the-art methods include ATOM \cite{danelljan2019atom}, DaSiamRPN \cite{zhu2018distractor} , DiMP\cite{bhat2019learning},  KYS \cite{bhat2020know}, SiamRPN++ \cite{li2019siamrpnpp}, SiamBAN \cite{chen2020siamese}, TrDiMP \cite{wang2021transformer}, SiamCAR \cite{guo2020siamcar} and TransT \cite{chen2021transformer}.
In Figure \ref{fig:otb}, our tracker delivers competitive performance compared to other methods. Particularly, our method achieves the highest success score among all compared methods in Figure \ref{fig:otb}(a). The result shows the effectiveness of our tracker in handling large appearance change, since the success score is computed based on the overlap rate of the predicted and ground truth bounding boxes.

\paragraph{GOT-10k}
The GOT-10k contains more than 10000 training sequences and 180 testing sequences of moving objects from the real-world. A unique property of GOT-10k is that the object classes between training and testing sets are zero-overlapped, which can show the tracker's ability of tracking unseen objects. We conduct the experiment on GOT-10k following the protocol that only GOT-10k training set is used for training our model. We report the average of overlap rates (AO), success rate $\text{SR}_{0.50}$ at threshold 0.5  and success rate $\text{SR}_{0.75}$  at threshold 0.75 
of our tracker and the state-of-the-art trackers including DiMP\cite{bhat2019learning}, Ocean \cite{zhang2020ocean}, KYS \cite{bhat2020know}, SiamFC++ \cite{SiamFC++}, PrDiMP \cite{danelljan2020probabilistic}, SiamR-CNN \cite{voigtlaender2020siam} and STARK-ST50 \cite{yan2021learning} in Table \ref{tab:got}. Our method achieves strong performance on GOT-10k comparing to other methods, which indicates the generalization ability of our method to track unseen objects. Although SiamR-CNN has a slightly higher AO than our method, our tracker is 6 times faster. 

Notably, the STARK-ST50 tracker outperforms our method by a significant margin. The performance gap might be caused by the following reasons: Firstly, STARK-ST50 trains a template scoring network to reject unreliable dynamic templates. while our method updates the target representation with the previous frame end-to-end. Though our update mechanism is simpler, it is more susceptible to occlusion and out-of-view challenges. Our dynamic target part representation can also be improved using this strategy, which we will investigate for our future works. Secondly, our tracker adopts a search region of 255 by 255 following the SiamRPN \cite{li2018high}, while that used in STARK-ST50 is 320 by 320. The larger search region may improve the tracker's robustness against fast motion and target our-of-view. 

\begin{table}[h]
\centering
\scalebox{0.8}{
\begin{tabular}{l c c c c c c c c c}
\hline
     & DiMP & Ocean  & SiamFC++ & KYS  &  PrDiMP  & SiamR-CNN  & STARK-ST50 & \textbf{Our} \\ 
     
   \hline
\textbf{$\text{SR}_{0.50}$}(\%)     & 71.7 &  72.1 & 69.5  & \textcolor{blue}{75.1} & 73.8 & 72.8 & \textcolor{red}{77.7} &\textcolor{green}{74.7}\\ 
\textbf{$\text{SR}_{0.75}$}(\%)    &  49.2 &  47.3  & 47.9 & 51.5 & 54.3 & \textcolor{blue}{59.7} & \textcolor{red}{62.3} & \textcolor{green}{54.6}\\ 
\textbf{AO}(\%)                  &  61.1 & 61.1 & 59.5 & 63.6 & 63.4 & \textcolor{blue}{64.9} & \textcolor{red}{68.0} & \textcolor{green}{64.2}\\ \hline
\textbf{fps}                &  - & - & - & - & - & 5 (V100) &30 (V100) & 33 (2080 ti)\\ \hline

\end{tabular}
}
\vspace*{0.5mm}
\caption{\label{tab:got} Comparison on GOT-10k with the state-of-the-art in terms of AO,  $\text{SR}_{0.50}$, $\text{SR}_{0.75}$ and fps. The top-3 results are shown in red, blue and green. DiMP denotes the ResNet-50 version (DiMP-50), and Ocean denotes the online version.
}
\end{table}

\section{Discussion}

\subsection{Component-based Analysis}

\paragraph{Effectiveness of Attention-guided Part Localization}
We validate the effectiveness of the attention-guided supervision imposed on the part location prediction by removing the attention loss during training. The evaluation results on VOT2018 and OTB100 denoted as \textbf{w/o atten. loss} are shown in Table \ref{tab:ablation}. The results show that the attention-guided  supervision significantly improves the tracking performance and part location prediction. Since the object bounding box is calculated based on the mean and standard deviations of the part locations, the bounding box penalty constrains the overall distribution of the target parts. However, individual part prediction lacks guidance and has too much flexibility. Thus constraining the part location prediction with their attended locations will force it to learn the semantic information and thereby it will predict more meaningful location.

\begin{table}[h]
\centering
\scalebox{0.9}{
\begin{tabular}{|l | c c c| c c|}
\hline
& \multicolumn{3}{c|}{VOT2018} &\multicolumn{2}{c|}{OTB100}    \\
   &\textbf{A}($\uparrow$)  &\textbf{R}($\downarrow$)   & \textbf{EAO}($\uparrow$) & \textbf{Suc.} & \textbf{Pre.} \\ 
   \hline
w/o atten. loss       &0.563   &0.302  & 0.322 & 0.674  & 0.871  \\
w/o updater    &0.604   &0.421   & 0.269  &0.675  & 0.873\\ 
our    & \textbf{0.617}   & \textbf{0.140}   & \textbf{0.489}  & \textbf{0.705} & \textbf{0.908} \\ 
\hline
\end{tabular}
}
\vspace*{0.5mm}
\caption{\label{tab:ablation} Evaluation results of variants of our method on VOT2018 and OTB100. 
}
\end{table}

\paragraph{Effectiveness of Dynamic Target Part Updating}
To show the effectiveness of the dynamic part representation updater, we train our network without the updater module and evaluate it on VOT2018 and OTB100. The evaluation results denoted as \textbf{w/o updater} are shown in Table \ref{tab:ablation}.  Without the dynamic target part representation, a large performance drop is observed. This indicates the dynamic target part representation module is crucial for adapting to object appearance change. The simple yet effective multi-head attention module learns to align pseudo template parts to the matched target parts and aggregate their representations to form the updated dynamic part representation.

\subsection{Qualitative Analysis}
\begin{figure}
    \centering
    \includegraphics[width=0.8\textwidth]{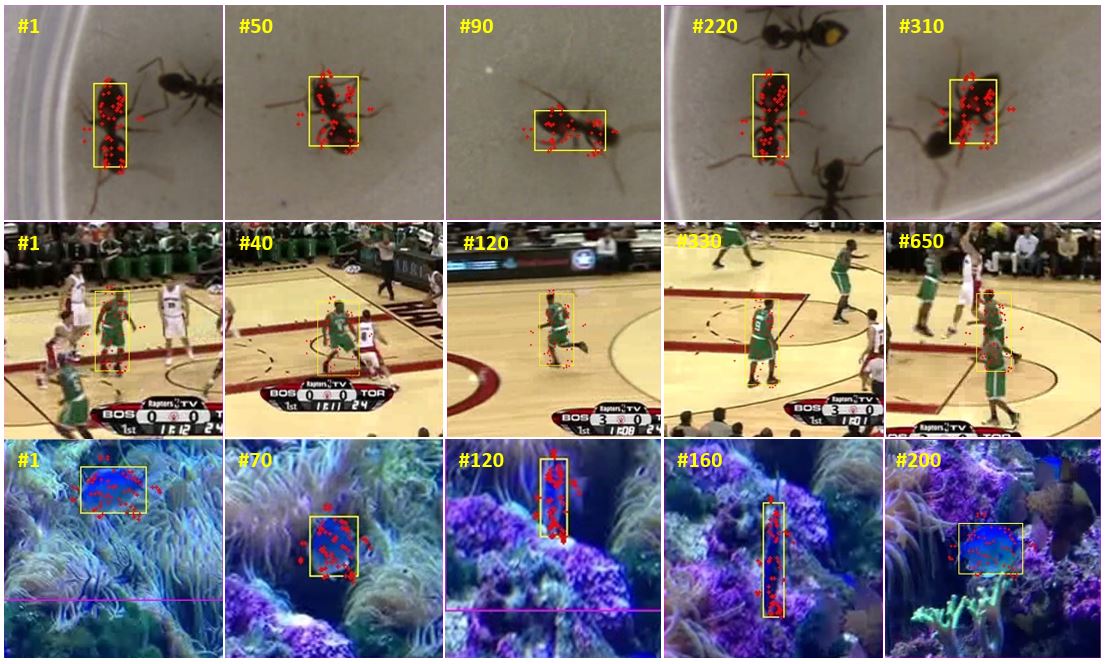}
 \caption{Visualization of the predicted target part centers depicted in red points and estimated bounding box on example frames from three representative sequences under challenging scenarios }\label{fig.vis}
\end{figure}

Since the ground truth part locations are unknown, we qualitatively analyze the part localization result by visualizing the predicted center points in Figure \ref{fig.vis}. We take three representative video sequences, namely \textit{ant1}, \textit{basketball} and \textit{fish1} from VOT2018 as examples. It demonstrates that most of the predicted part centers coincide well with the object silhouette even in challenging scenarios such as large object deformation (\textit{fish1},\textit{basketball}), object rotation (\textit{ant1}) and distractor (\textit{ant1}, \textit{basketball}). Attributed to the accurate and flexible part localization, our tracker can estimate the bounding box precisely. We notice that some predicted part center points are located in the background, which might result from the rectangular bounding box we use to identify target parts. Since most objects are of irregular shapes, some part representation vectors in the feature map indicated as target parts actually belong to the background, which lead to predicted locations outside the target object contour.

\section{Conclusion}
In this work, we propose a dynamic part-based visual tracker to online update the target part representation. Moreover, we design an attention-guided part localization network to directly predict the part localization. The final bounding box is further determined by the distribution of part localization. The whole method is an end-to-end with a simple yet effective transformer encoder.  Our proposed tracker can achieve state-of-the-art or comparable results on various benchmarks: VOT2018 \cite{kristan2018sixth}, OTB100 \cite{WuLY15} and GOT-10k \cite{huang2019got}.

\bibliography{egbib}
\end{document}